\documentclass{article} 
\usepackage{iclr2026_conference,times}

\usepackage{amsmath,amsfonts,bm}

\def\eqref#1{equation~\ref{#1}}

\def\1{\bm{1}}

\DeclareMathAlphabet{\mathsfit}{\encodingdefault}{\sfdefault}{m}{sl}
\SetMathAlphabet{\mathsfit}{bold}{\encodingdefault}{\sfdefault}{bx}{n}

\usepackage{hyperref}
\usepackage{url}
\definecolor{iclrCite}{HTML}{0A58CA}
\hypersetup{
    colorlinks=true,
    linkcolor=black,
    citecolor=iclrCite,
    urlcolor=iclrCite
}

\let\cite\citep

\iclrsetauthors{Yushuo Zheng$^{12}$, Tongrui Ye$^{3}$, Zicheng Zhang$^{12}$,\\
Xiongkuo Min$^{1}$, Huiyu Duan$^{1}$\thanks{Corresponding authors.}, Guangtao Zhai$^{12}$\footnotemark[1]}{%
$^{1}$ Shanghai Jiao Tong University, Shanghai, China\\
$^{2}$ Shanghai AI Lab, Shanghai, China\\
$^{3}$ Tongji University, Shanghai, China\\
\texttt{\{yushuozheng, zzc1998, minxiongkuo, huiyuduan, zhaiguangtao\}@sjtu.edu.cn}\\
\texttt{2433480@tongji.edu.cn}
}
\author{}

\iclrfinalcopy 
\begin{document}

\title{LM Fight Arena: Benchmarking Large Multimodal Models via Game Competition}

\maketitle
\begin{figure*}[ht]
   \centering
    \includegraphics[width=\textwidth]{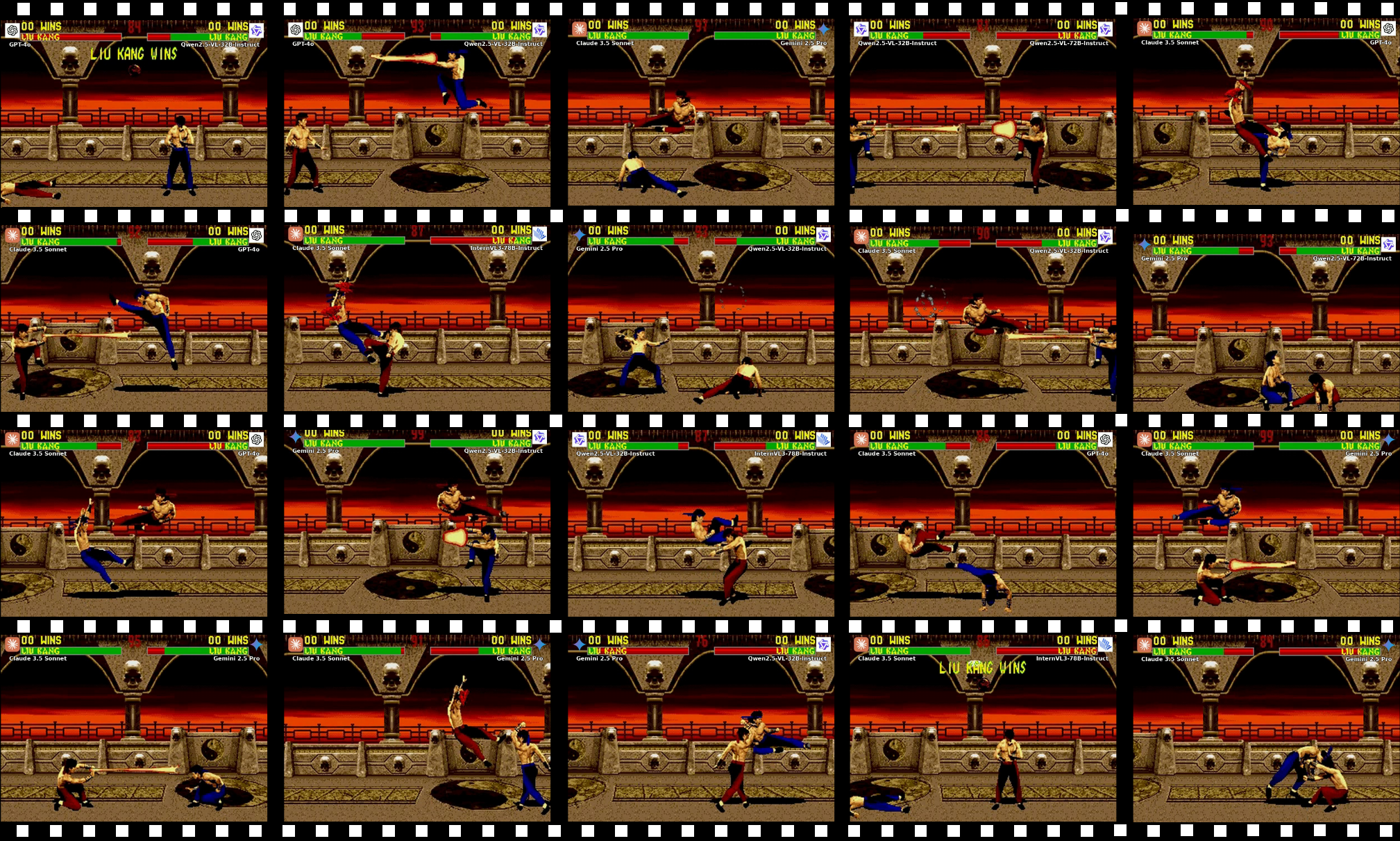} 
    \caption{The LM Fight Arena. Six state-of-the-art large multimodal models (LMMs) compete in a round-robin tournament in the classic fighting game \textit{Mortal Kombat II}. Each model controls the same character, Liu Kang, to ensure a fair comparison. The models receive real-time visual frames and structured game state information, then output their next actions as natural language commands.}
\end{figure*}
\begin{abstract}
Existing benchmarks for large multimodal models (LMMs) often fail to capture their performance in real-time, adversarial environments. We introduce \textbf{LM Fight Arena (Large Model Fight Arena)}, a novel framework that evaluates LMMs by pitting them against each other in the classic fighting game \textit{Mortal Kombat II}, a task requiring rapid visual understanding and tactical, sequential decision-making. In a controlled tournament, we test six leading open- and closed-source models, where each agent operates controlling the same character to ensure a fair comparison. The models are prompted to interpret game frames and state data to select their next actions. Unlike static evaluations, LM Fight Arena provides a fully automated, reproducible, and objective assessment of an LMM's strategic reasoning capabilities in a dynamic setting. This work introduces a challenging and engaging benchmark that bridges the gap between AI evaluation and interactive entertainment.
\end{abstract}


\section{Introduction}

The rapid advancement of large multimodal models (LMMs), which integrate visual perception with language understanding, has marked a significant milestone in artificial intelligence \cite{flamingo,gpt4,claude3}. While these models excel at tasks like image captioning and visual question answering \cite{llava,blip2}, evaluating their true capabilities remains a challenge. Traditional benchmarks, often based on static image-text pairs or curated datasets, do not capture the dynamic, interactive nature of real-world applications where agents must perceive, reason, and act in real-time \cite{mmmu,mmbench}.

To address this gap, we turn to fighting games, a uniquely challenging domain for AI evaluation. Unlike turn-based strategy games such as chess or Go, fighting games operate at a frame-level granularity, demanding immediate visual processing and low-latency decision-making. Success requires sophisticated spatial and temporal reasoning to track character positions, predict opponent moves, and execute precise combos. This adversarial environment compels models to balance offense with defense and continually adapt their strategy, making it an ideal testbed for assessing multimodal reasoning in a dynamic context \cite{dqn,alphago}.

\textit{Mortal Kombat II} (1993) is selected as our evaluation environment because it offers a compelling balance of complexity and interpretability. The game's structured mechanics, including clear health bars and distinct character animations, provide reliable visual information. Its action space, governed by an 8-button controller, allows for rich strategic possibilities. Furthermore, the visual game states are easily understood, enabling qualitative analysis of model behavior.

This paper introduces \textbf{LM Fight Arena}, the first fully automated framework for evaluating LMMs in a fighting game environment. Our primary contributions include a comprehensive evaluation of six leading models in direct adversarial matchups and an experimental design that ensures fairness by having all models control the same character. We provide a reproducible methodology and objective evaluation metrics that offer clear, quantitative assessments of performance. In doing so, we establish fighting games as a valuable new direction for L-MM evaluation, complementing existing benchmarks.

\section{Related Work}
\label{sec:related}

The evaluation of large multimodal models has evolved significantly from early benchmarks focused on static tasks like image captioning and visual question answering (VQA), such as VQAv2, COCO Captions, and Flickr30k \cite{vqav2, coco, flickr30k}. While foundational, these datasets primarily assess static image understanding. More recent benchmarks have expanded this scope, with frameworks like MMBench, MME, and MMMU testing a wider range of perception and reasoning skills across various disciplines \cite{mmbench, mme, mmmu}. Others, such as LLaVA-Bench, have focused on evaluating instruction-following capabilities in multimodal contexts \cite{llava}. The field has also moved towards assessing performance on interactive and sequential tasks through environments like VisualWebArena for web navigation and AndroidWorld for mobile app interaction \cite{visualwebarena, androidworld, mobileenv}. The concept of a competitive evaluation was notably pioneered by the Chatbot Arena, which uses pairwise human preference to rank models, inspiring our automated, adversarial approach \cite{chatbotarena}.

Our work also builds on the rich history of game-playing AI. Milestones in this field include Deep Blue's mastery of chess and AlphaGo's success in Go, both of which relied on deep search in turn-based environments \cite{deepblue, alphago}. More recently, deep reinforcement learning has enabled agents to achieve human-level performance in real-time video games, from classic Atari titles to complex strategy games like StarCraft II and Dota 2 \cite{dqn, alphastar, openai5}. Research in fighting game AI has traditionally used methods like finite state machines or deep reinforcement learning, but these systems require extensive, game-specific training and do not test the zero-shot reasoning capabilities of general-purpose models \cite{fightingai1, fightingai2, fightingai3}.

The application of vision-language models (VLMs) to gaming is an emerging area of research. While models have shown promise in understanding game UIs and providing strategic advice from static screenshots \cite{vlmgaming1, vlmgaming2, gpt4vision}, their ability to perform in dynamic, interactive settings is less understood. Our work evaluates the state-of-the-art in this domain, including powerful open-source models like InternVL3 and the Qwen2-VL series, which have demonstrated strong performance on a variety of vision-language benchmarks \cite{internvl, qwen2vl, qwen25vl}.

Despite these advances, a significant gap remains in the literature. Existing benchmarks do not subject LMMs to real-time, adversarial competition where frame-level visual processing and immediate action are required for success. LM Fight Arena addresses this gap by providing a fully automated and interpretable environment that directly evaluates the sequential decision-making of LMMs in a competitive setting.

\section{Methodology}
\label{sec:methodology}
\begin{figure*}
    \centering
    \includegraphics[width=\textwidth]{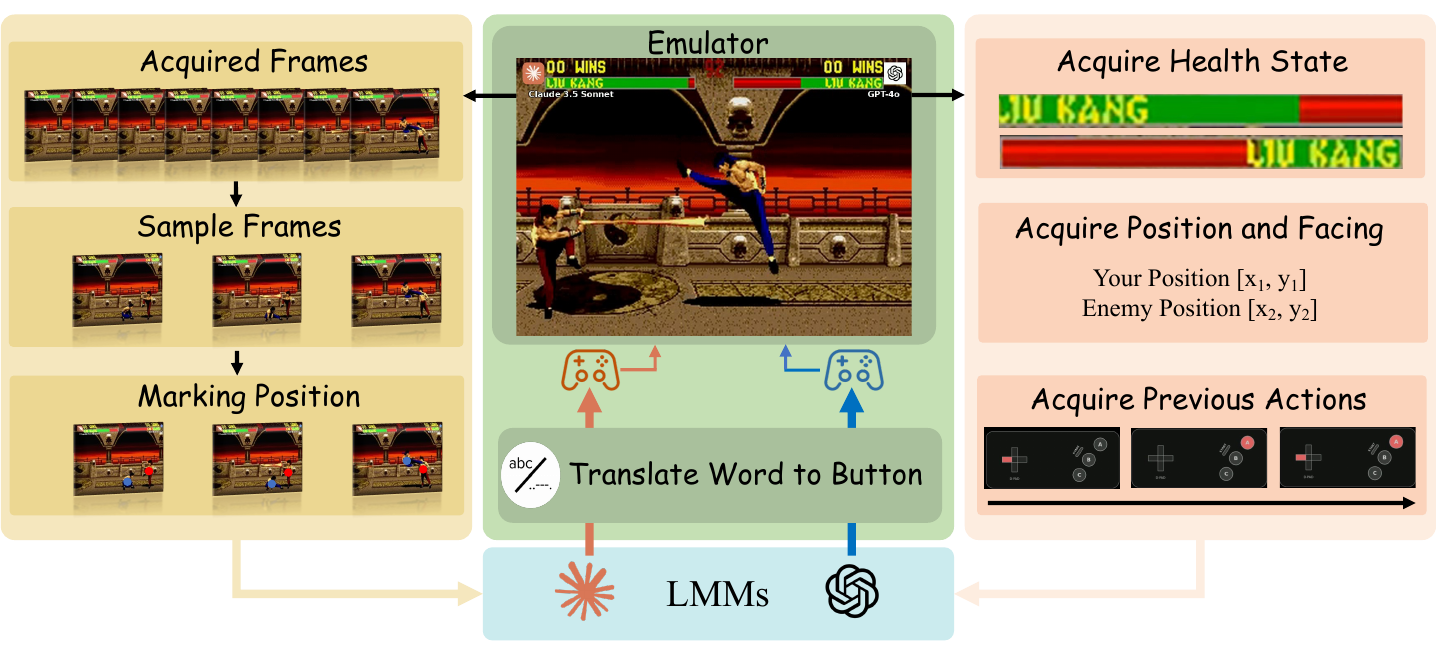}
    \caption{Overview of the LM Fight Arena control loop. The left column illustrates visual processing: the emulator streams raw frames, we subsample every fourth frame, and we annotate player positions before packaging the sequence. The central block shows the real-time interaction between the emulator and the language-driven controller that translates text actions into Sega Genesis button presses. The right column depicts the structured state features---health bars, absolute coordinates, facing direction, and the trailing action history---that accompany the visual stack.}
    \label{fig:methodology}
\end{figure*}
The evaluation framework is built around a carefully controlled and reproducible environment (Figure~\ref{fig:methodology}). \textit{Mortal Kombat II} (Sega Genesis version, 1994) is selected as the testbed. This setup provides deterministic game states and lossless frame capture, which are essential for reproducible experiments. To ensure a fair and direct comparison of model capabilities, all matches are mirror matches where both competing agents control the character Liu Kang. This design eliminates any character-specific advantages, allowing us to isolate each model's performance in perception, reasoning, and decision-making.

Figure~\ref{fig:methodology} highlights the multimodal observation pipeline. The emulator continuously obtains 10 frames at 4-frame intervals (approximately 1 seconds of context). We overlay player-specific markers to disambiguate positions before base64-encoding the sampled frames. In parallel, we read the game's memory to extract structured features: current health bars, $(x,y)$ coordinates, facing direction, and the last five executed actions. These signals are bundled into a natural-language state description so that every agent receives an aligned mixture of visual and symbolic cues.

The action space consists of all button combinations on the Sega Genesis controller. Each model returns a natural-language command (e.g., ''Left + A'' or ''Down, Forward, A''). The ''Translate Word to Button'' module shown in Figure~\ref{fig:methodology} parses these strings, resolves timing-sensitive sequences, and injects the corresponding button presses into the emulator for execution. All models conduct their control for each frame.

We evaluate six leading multimodal models, including three open-source (InternVL3-78B-Instruct, Qwen2.5-VL-32B-Instruct, and Qwen2.5-VL-72B-Instruct) and three closed-source (Claude-3.5-Sonnet, Gemini-2.5-Pro, and GPT-4o) systems. All models are given an identical system prompt that outlines the game's objective, the observation format, and the available actions, ensuring a consistent starting point for all competitors.

The evaluation takes the form of a round-robin tournament where each model faces every other model once. Performance is measured using two key metrics: the primary metric is the binary win/loss record, while a secondary metric of the winner's remaining health percentage provides a more granular assessment of performance dominance. The results are presented in a diagonal tournament matrix, which allows for a clear visualization of each model's overall standing and its performance in specific matchups.

To ensure scientific rigor, we have taken several steps to promote reproducibility, including using deterministic emulation with fixed random seeds and documenting all model versions. However, we acknowledge several limitations in the current study, such as conducting only a single match per pair, and the fact that all models are evaluated in a zero-shot setting without any game-specific fine-tuning.

\section{Results}
\label{sec:results}

\begin{figure*}
    \centering
    \includegraphics[width=\textwidth]{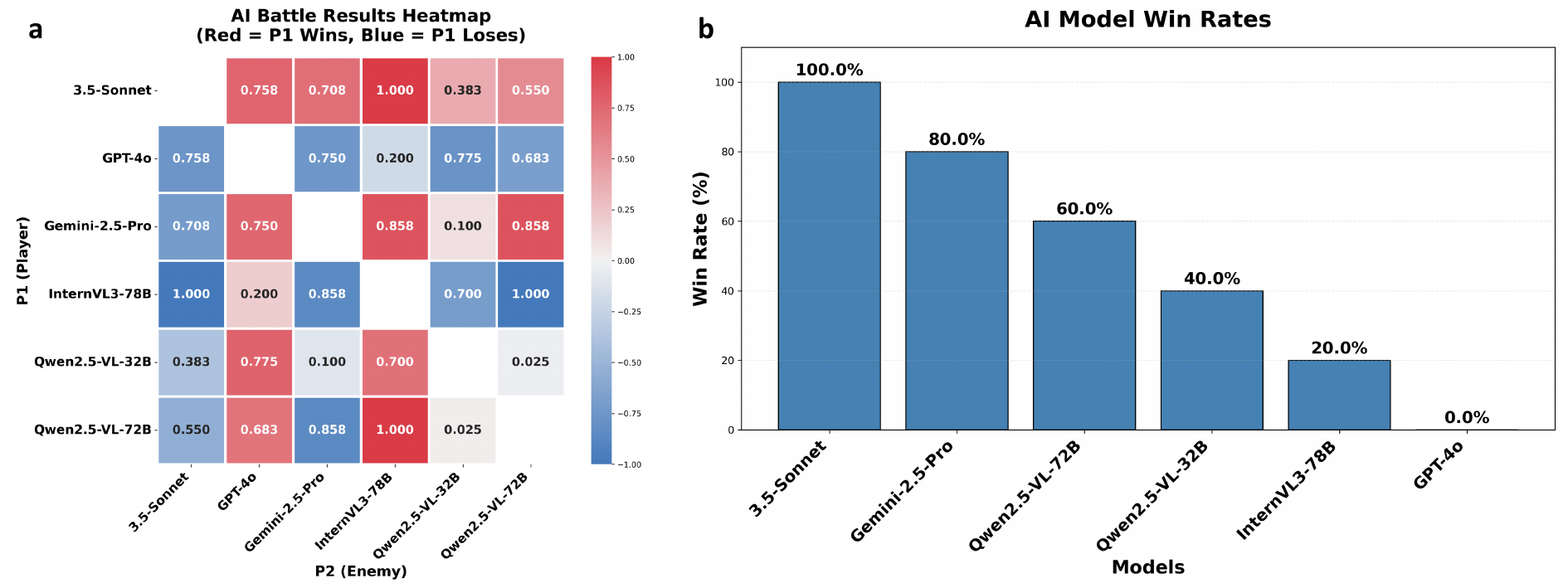}
    \caption{(a) Tournament matchup matrix showing the winner's remaining health percentage for each model pair. Green cells indicate a win for the row model, red cells indicate a loss. (b) Bar chart summarizing overall win rates for each model.}
    \label{fig:fight_result}
\end{figure*}

The LM Fight Arena tournament highlights a clear stratification among the six evaluated large multimodal models (Figure~\ref{fig:fight_result}). Claude~3.5~Sonnet completed the round robin undefeated, securing first place with a $100\%$ win rate. Gemini~2.5~Pro followed with $80\%$, powered by decisive victories over every open-source model and a narrow loss to Claude. The Qwen family split the middle of the leaderboard: the larger Qwen2.5-VL-72B achieved $60\%$ by consistently exploiting GPT-4o's defensive lapses and edging out InternVL3-78B, while Qwen2.5-VL-32B landed in fourth at $40\%$. InternVL3-78B struggled offensively and finished at $20\%$, and GPT-4o failed to register a win despite frequent counter-punch attempts.
\begin{figure*}
    \centering
    \includegraphics[width=\textwidth]{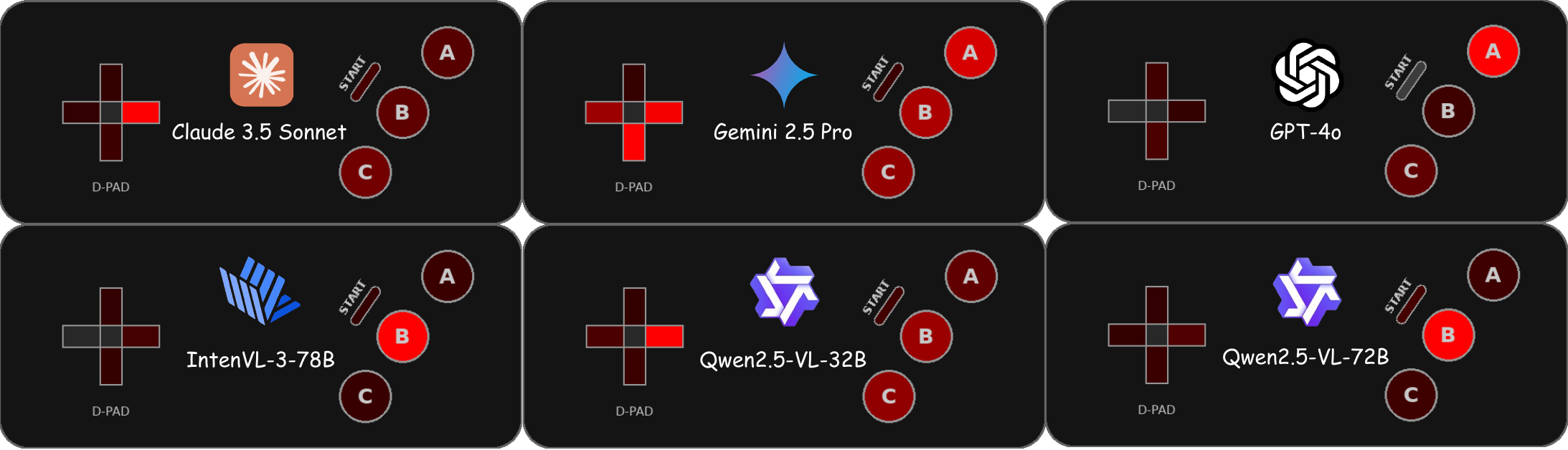}
    \caption{Heatmap of button press frequencies across all models and matches. Brighter colors indicate more frequent usage of that button.}
    \label{fig:controller_heat}
\end{figure*}

\subsection{Overall Tournament Results}
The matchup matrix in Figure~\ref{fig:fight_result}a encodes each bout as the winner's remaining health percentage. Claude~3.5~Sonnet recorded decisive $0.758$--$1.000$ health margins against every opponent. Gemini~2.5~Pro posted similarly dominant red cells except for a $0.200$ loss to Claude. Qwen2.5-VL-72B exhibited the most polarized performance: it demolished InternVL3-78B ($1.000$) yet barely survived Qwen2.5-VL-32B ($0.025$). The lower-triangular blue cells reveal GPT-4o's vulnerability---it ceded $0.775$ health to Claude and $0.683$ to Qwen2.5-VL-72B. Aggregating across rows yields the bar chart in Figure~\ref{fig:fight_result}b, where the win-rate ordering mirrors the qualitative impressions from the heatmap.

\subsection{Model Performance Analysis}
Closed-source systems outperformed open-source counterparts by a wide margin. Claude and Gemini combined for 11 of the 15 match wins and averaged $0.66$ remaining health in their victories. Among open models, Qwen2.5-VL-72B proved the most resilient, using longer button strings to punish passive opponents. Qwen2.5-VL-32B relied on single-button pokes and paid a health tax when fights extended past mid-clock. InternVL3-78B's $0.200$ win margin over GPT-4o was its lone bright spot, and it absorbed the only $1.000$-health defeat delivered by an open-source rival.

\subsection{Qualitative Observations}
Controller heatmaps in \ref{fig:controller_heat} expose how strategies translated to inputs. Claude and Gemini saturated every spoke of the directional pad while lighting up the ``B'' button, evidence of relentless forward dashes that immediately transitioned into blocks or counter-sweeps. Gemini also showed brighter ``A'' usage than Claude, aligning with its preference for jump-in punches before executing the ``Forward, Forward, C'' flying kick. GPT-4o, by contrast, rendered the D-pad almost gray and concentrated activity on a single ``A'' button, confirming that the model often stayed rooted in place throwing high punches until it was cornered. InternVL3-78B's controller shows a similar dependence on ``B'' without the accompanying directional inputs required for cancels, explaining why its pressure fizzled into idle blocking. The Qwen pair occupied the middle ground: both panels display alternating forward and backward taps, but Qwen2.5-VL-72B saturates the ``B'' button and the down vector, reflecting crouch–kick zoning, whereas the 32B model exhibits more even shading consistent with tentative feints. Across all agents we observed limited adaptation once a sequence began repeating, underscoring the challenge of on-the-fly strategic reasoning.

\section{Discussion}
\label{sec:discussion}

The LM Fight Arena results highlight how architectural priors and latency constraints translate into competitive performance. Claude~3.5~Sonnet's undefeated run stems from rapid intent generation and strong multimodal grounding: the model consistently fused visual and textual cues to trigger high-damage combos, suggesting its training emphasizes instruction-following under tight temporal windows. Gemini~2.5~Pro displayed similarly polished control, though its single loss indicates occasional hesitation when forced to defend against unconventional strings. In contrast, GPT-4o's zero-win record points to a policy that over-indexes on safe responses. Even with accurate perception, response delays pushed it into passive loops that opponents exploited. Among open-source models, Qwen2.5-VL-72B capitalized on its larger context handling to set up zoning patterns, whereas the 32B variant and InternVL3-78B oscillated between aggressive bursts and prolonged blocking, reflecting limited capacity to sustain coherent plans.

\subsection{Insights into Multimodal Reasoning}
Successful LMMs paired precise visual parsing with rapid temporal reasoning. Claude and Gemini demonstrated reliable tracking of opponent states, adjusting button sequences when the adversary stood, crouched, or jumped. The heatmap evidence shows they coordinated dashes and blocks, implying they reasoned over short action histories to predict reversals. Qwen2.5-VL-72B's zoning revolves around recognizing spacing advantages, but its narrow victory over Qwen2.5-VL-32B underscores the challenge of adapting once the opponent mirrors defensive postures. GPT-4o can not perceive threats with zero block output across the tournament. These behaviors suggest current LMMs can parse frame-level signals but still struggle to plan multi-step contingencies.

\subsection{Comparison with Existing Benchmarks}
Traditional VQA or captioning benchmarks reward static recognition, whereas LM Fight Arena stresses continuous control under adversarial pressure. The divergence between GPT-4o's strong performance on static tasks and its tournament collapse illustrates this gap: high linguistic competence does not guarantee effective closed-loop decision-making. Conversely, Claude's dominance aligns with reports of its fast tool-use abilities, hinting that latency-aware training transfers to game-playing. This benchmark therefore uncovers facets of multimodal reasoning---anticipation, adaptation, and timing---that remain underexamined in existing datasets.

\subsection{Limitations}
Our evaluation still faces limitations. Single matches per pairing restrict statistical confidence; extending to larger series would mitigate variance in close contests such as the Qwen mirror. Prompt formatting and controller parsing may bias models toward conservative behavior, particularly GPT-4o. Closed-source APIs introduce latency variability that we cannot fully control, and no model received fine-tuning or few-shot demonstrations that might unlock richer strategies. Finally, Mortal Kombat II with Liu Kang captures only one genre and character archetype, leaving open how results generalize to other games or control schemes.

\subsection{Broader Implications}
Despite these constraints, the benchmark suggests practical implications. Fighting games provide a controlled yet dynamic playground for assessing real-time decision quality, making LM Fight Arena a candidate for standardized stress testing before deployment of embodied assistants. The techniques for synchronizing multimodal observations with language-based action plans can transfer to robotics, teleoperation, or human-AI teaming. Ethical considerations also emerge: competitive scenarios reward aggressive behaviors, so safeguards are needed before transferring such policies to collaborative contexts.

\subsection{Future Work}

Future research should first reinforce statistical robustness by expanding each pairing to multi-match series (10--100 bouts) so that narrow margins such as the Qwen mirror can be interpreted with confidence intervals. Extending the arena to other franchises---for example Street Fighter II with its six-button inputs, Tekken 3 with sidestepping, or the team battles of King of Fighters---would expose models to richer mechanics and test whether strategies transfer beyond Liu Kang's move set. Similar diversity is needed at the character level within Mortal Kombat II; forcing agents to adapt to grapplers or zoners would reveal whether they can internalize counter-pick logic rather than memorizing a single matchup.

Another direction is to explore in-context learning and fine-tuning. Providing example transcripts or strategy guides before competition could measure how rapidly models absorb fighting concepts, while targeted fine-tuning on annotated match footage would clarify the ceiling achievable through domain adaptation and the data efficiency required for specialized control. Human calibration matches remain essential for grounding the benchmark: exhibitions against experienced players would expose superhuman bursts, highlight subhuman blind spots, and generate interpretable commentary on model decision processes.

\section{Conclusion}
\label{sec:conclusion}

We introduced \textbf{LM Fight Arena (Large Model Fight Arena)}, a novel benchmark framework for evaluating large multimodal models through adversarial fighting game competition. By leveraging Mortal Kombat II as a controlled testbed, we provide an automated, reproducible, and interpretable method for assessing real-time visual perception, sequential decision-making, and strategic reasoning capabilities of state-of-the-art LMMs.

Our framework addresses critical gaps in existing evaluation methodologies by enabling direct adversarial competition between models without human intervention, testing frame-level visual comprehension alongside immediate action selection, providing interpretable game environments for qualitative analysis, and offering objective metrics through win/loss records and health percentages.

The tournament-style evaluation of six leading models---InternVL3-78B, Qwen2.5-VL-32B/72B, Claude-3.5-Sonnet, Gemini-2.5-Pro, and GPT-4o---reveals a pronounced performance hierarchy. Claude~3.5~Sonnet achieved a perfect record and the largest health margins, followed by Gemini~2.5~Pro at an $80\%$ win rate and Qwen2.5-VL-72B at $60\%$. Qwen2.5-VL-32B and InternVL3-78B displayed intermittent success, while GPT-4o failed to secure a win, struggling to convert its perceptual understanding into timely actions. These outcomes demonstrate the potential of fighting games as challenging, engaging benchmarks for multimodal AI and emphasize the importance of tight perception-action coupling.

As LMMs continue to advance toward general-purpose visual intelligence, evaluation frameworks must evolve beyond static datasets to embrace dynamic, interactive scenarios. LM Fight Arena represents a step in this direction, establishing fighting games as a valuable domain for rigorous, reproducible, and entertaining AI assessment.

\textbf{We hope LM Fight Arena inspires future research at the intersection of multimodal AI, game playing, and competitive evaluation frameworks.}


\bibliography{iclr2026_conference}
\bibliographystyle{iclr2026_conference}

\end{document}